\documentclass[runningheads]{llncs}

\usepackage{eccv}

\usepackage[colorlinks=true, linkcolor=blue, urlcolor=blue]{hyperref}
\usepackage{eccvabbrv}
\usepackage{enumitem}
\usepackage{graphicx}
\usepackage{booktabs}
\usepackage{diagbox}

\usepackage[accsupp]{axessibility}  

\usepackage{hyperref}
\usepackage[table,xcdraw]{xcolor}
\usepackage{authblk}
\definecolor{top1color}{RGB}{255, 180, 180}   
\definecolor{top2color}{RGB}{220, 235, 255}
\usepackage{orcidlink}
\usepackage{adjustbox}

\begin{document}

\title{P²Fusion: Prompt-based Progressive Infrared-Visible Image Fusion via Dual-Prior Distillation} 


\author{
    Yi Shi\inst{1}\textsuperscript{$\dagger$} \and 
    Huichao Xie\inst{1}\textsuperscript{$\dagger$} \and 
    Yuqing Wang\inst{1} \and 
    Mingyu Wang\inst{1} \and 
    Kaihui Yang\inst{1} \and 
    Yu Liu\inst{2} \and 
    Ruitao Lu\inst{3} \and 
    Lizhe Li\inst{1} \and 
    Junwei Han\inst{1,4}\textsuperscript{*} \and 
    Dingwen Zhang\inst{1}\textsuperscript{*}
}

\authorrunning{Y. Shi et al.}

\institute{
Northwestern Polytechnical University, Xi'an, China \and
Hefei University of Technology, Hefei, China \and
Rocket Force University of Engineering, Xi'an, China \and
Chongqing University of Posts and Telecommunications, Chongqing, China
}

\maketitle

\footnotetext[1]{\textsuperscript{$\dagger$} These authors contributed equally to this work.}
\footnotetext[2]{\textsuperscript{*} Corresponding authors: Junwei Han (jhan@nwpu.edu.cn), Dingwen Zhang (zdw2006yyy@nwpu.edu.cn).}

\begin{abstract}
Infrared-visible image fusion (IVIF) is pivotal for multimodal perception, yet reconciling the inherent information disparity between thermal and textural features remains a fundamental challenge. Existing prior-guided methods often rely on static constraints that induce optimization conflicts or utilize extrinsic semantic priors from large-scale foundation models (e.g., CLIP/DINO), which frequently fail to exploit the intrinsic modality characteristics essential for high-fidelity fusion. To address these issues, we propose P²Fusion, a prior-guided distillation-based framework that reformulates IVIF via dual intrinsic prompts. Instead of imposing hard-coded penalties, we distill image-intrinsic priors, thermal saliency and spatial quality—into learnable, dynamic regulators. Specifically, a Teach-to-Fuse mechanism provides dual-granularity progressive guidance, coupled with a Gated Dynamic Expert Recalibration (GDER) module for decoupled feature refinement. This design enables the network to adaptively mediate modal competition through expert specialization. Extensive experiments demonstrate that P²Fusion achieves state-of-the-art performance across five mainstream datasets.  Notably, our framework demonstrates consistent performance advantages in fusion quality, achieving state-of-the-art results in 14 out of 20 key evaluation metrics across 5 benchmarks. Furthermore, it effectively contributes to the robustness of downstream perception, such as +3.2\% mAP on MSRS, +0.5\% mAP on M3FD and +0.9\% mAP on DroneVehicle for object detection. Our code will be available at \url{https://github.com/YiShi99/P2Fusion}.
  \keywords{ Infrared-visible image fusion \and Prompt-based learning \and Knowledge distillation \and Multimodal perception}
\end{abstract}

\begin{figure*} 
  \centering
  \includegraphics[width=\linewidth]{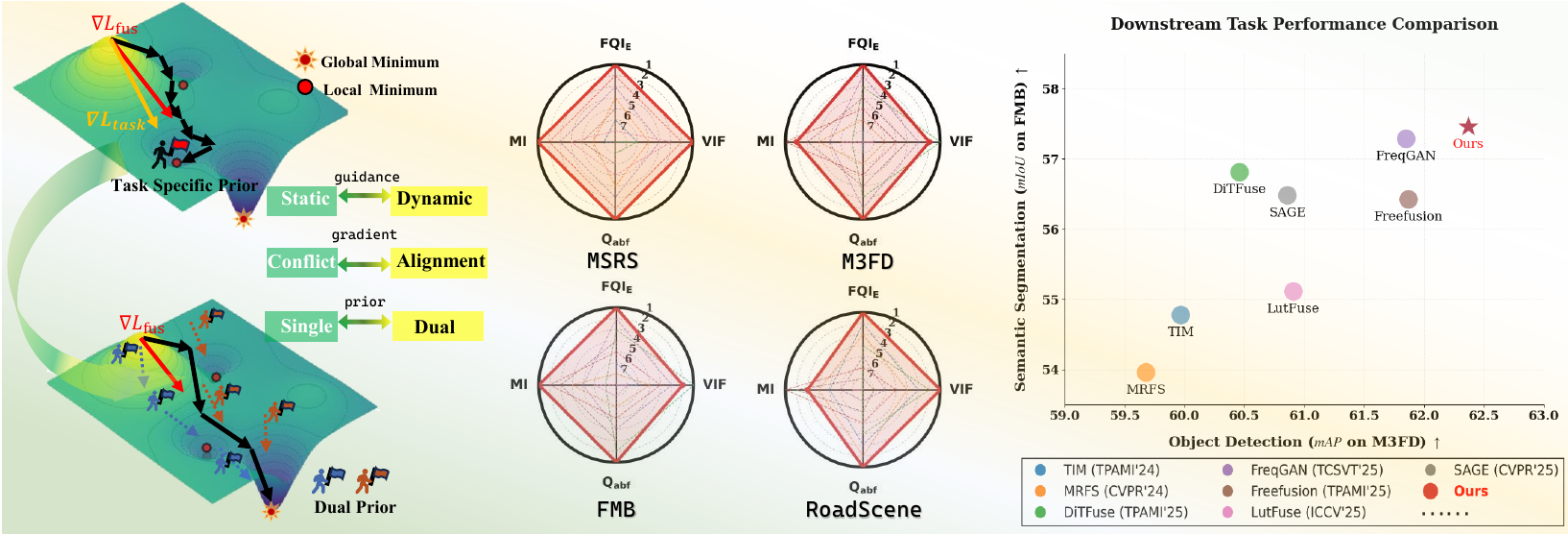} 
  \caption{\textbf{Motivation and Performance Overview.} (a) Prior integration paradigm (Left): Existing methods inject priors as static constraints, often causing optimization conflicts. Our P\textsuperscript{2}Fusion provides dynamic, prompt-based guidance via image-intrinsic priors to mitigate gradient interference. (b) Fusion quality (Middle): Radar charts across four datasets (MSRS, M3FD, FMB, RoadScene) demonstrate our superior visual fidelity. (c) Downstream perception (Right): P\textsuperscript{2}Fusion consistently enhances object detection and semantic segmentation performance on FMB and M3FD benchmarks.}
  \label{fig:fig1}
\end{figure*}

\section{Introduction}
\label{sec:intro}
Multi-modal image fusion aims to integrate complementary information from diverse sensors to produce a comprehensive representation with rich details~\cite{segmif,text-if}. This paradigm is indispensable in critical applications such as autonomous driving~\cite{auto-drive1,MFNet,Gao2026CoSurfGS}, target re-identification~\cite{identification1,identification}, and medical imaging~\cite{cdd}. Among these, Infrared–Visible Image Fusion (IVIF) has emerged as a predominant research direction, leveraging the inherent synergy between thermal radiation and textural details~\cite{tardal,swinfusion}. Since IVIF is essentially an unsupervised generative task due to the lack of paired ground-truth images, its performance heavily relies on the heuristic design of network architectures and loss functions~\cite{SHIP,FreeFusion}. Without explicit labels, existing methods often struggle to balance detail fidelity and structural integrity, leading to the emergence of prior-guided fusion methods~\cite{STDFusionNet,PIAFusion,SAGE,DCEvo}. By incorporating modality-specific cues, these methods provide indirect supervision to endow networks with scene-adaptive capabilities.

However, we observe that existing prior-guided paradigms face fundamental limitations across three mainstream technical routes. Firstly, methods based on explicit modality-specific hard-constraints~\cite{STDFusionNet,PIAFusion,CoConet} often treat segmentation masks or saliency maps as static regularization terms. This forces the network to prioritize prior distribution fitting over pixel-level feature integration, compelling the model to sacrifice fine-grained textures to satisfy rigid prior boundaries. Secondly, the route of joint optimization with downstream tasks~\cite{tardal,SAGE,DCEvo,Li2026LangSurf} integrates object detection or segmentation as auxiliary tasks. Nevertheless, these frameworks frequently introduce multi-objective optimization conflicts, where the drive for task-level accuracy compromises background detail preservation, creating a persistent trade-off between visual aesthetics and task performance. Thirdly, the recent trend of extrinsic semantic guidance~\cite{SPDFusion,text-if} leverages foundation models (e.g., CLIP or DINO) as soft guidance. While mitigating the "hard-constraint" issue, it introduces extrinsic information that deviates from the modal essence. Since image fusion is fundamentally a pixel-level task, the abstract high-level representations lead to a granularity mismatch between semantics and pixels, preventing precise guidance for low-level textural reconstruction.
 
Based on these insights, we propose a novel Teach-to-Fuse (T2F) paradigm (Fig.~\ref{fig:fig1}) that returns to the essence of image data. Following this paradigm, we develop \text{P\textsuperscript{2}Fusion}, a network built on the core philosophy that priors should be endogenous and dynamic. Rather than introducing extrinsic or rigid constraints, we focus on maximizing the mining of intrinsic attributes from the infrared and visible images themselves. We move away from seeking a "perfect" static prior and instead build an adaptive prompt learning framework. In this work, we select infrared thermal saliency and visible spatial quality as two complementary intrinsic priors. These two priors are chosen because they inherently capture distinctive modality traits and intuitively align with human visual perception, as supported by established literature \cite{CoConet}~\cite{PIAFusion}. Crucially, our framework serves to instantiate a prior-guided prompt-based philosophy rather than chasing the "best" theoretical teachers. It is prior-agnostic and can seamlessly extend to other task-related auxiliary cues (e.g., depth, semantics), thereby releasing optimization freedom and achieving dynamic scene adaptation.

To further optimize fusion performance and strengthen cross-modal interactions, we design a fusion architecture featuring a Gated Dynamic Expert Recalibration (GDER) module, which functions as a self-correcting feedback mechanism. Unlike traditional Mixture-of-Experts (MoE) used for capacity expansion, our GDER module is tasked with feature decoupling and correction. Within this module, we synergize a prior-guided Modality Specialist with a global-aware Attention Specialist to perform modality-specific feedback refinement. By leveraging a reliability-based gating mechanism to adaptively recalibrate the fusion strategy, we effectively decouple conflicting features and rectify biased signals, thereby ensuring a robust and perceptually consistent fusion representation.

The main contributions of this work are summarized as follows:

\begin{itemize}[topsep=0pt, partopsep=0pt, itemsep=2pt, parsep=0pt]
    \item We propose the \textbf{"Teach-to-Fuse"} paradigm for multi-modal image fusion. By systematically addressing the pitfalls of hard-constrained priors, multi-objective conflicts, and granularity mismatch, we transform static penalties into process-oriented guidance that maximizes the mining of intrinsic modality attributes via dynamic prompts.
    
    \item We develop the \textbf{Gated Dynamic Expert Recalibration (GDER)} module to optimize feature refinement and cross-modal interactions. This MoE-based self-correction mechanism synergizes a prior-guided Modality Specialist with a global-aware Attention Specialist, enabling the network to adaptively decouple conflicting features and rectify biased signals through modality-specific feedback.
    
    \item \textbf{Extensive Benchmarking and Perceptual Analysis.} We conduct a comprehensive evaluation of the proposed framework across five public datasets and over ten benchmark tasks. The results consistently demonstrate the superiority of our Teach-to-Fuse paradigm in terms of high-fidelity visual fusion, robust cross-domain generalization, and significant performance gains in downstream perception tasks.
\end{itemize}

\section{Related Work}
\label{related work}

\subsection{Deep Learning Paradigms in IVIF}
Deep learning has become the prevailing paradigm in IVIF, outperforming conventional methodologies such as sparse representation and multiscale decomposition~\cite{Gradient-based,decompositions,Sparse,ICA}. Early deep fusion frameworks primarily explored diverse architectures to improve feature representation, including Auto-Encoders (AEs)~\cite{Hierarchically,DenseFuse,RFN}, CNNs~\cite{review,Edge-Attention,emma}, Generative Adversarial Networks (GANs)~\cite{GANMcC,Coupled-GAN, MEF-GAN}, and more recently, Transformers~\cite{swinfusion,YDTR,cdd} and Diffusion Models~\cite{DDFM,Mask-DiFuser}.

These advanced architectures have demonstrated notable advantages in synthesizing complementary information through attention mechanisms~\cite{MetaFusion,DATFuse}, sophisticated feature interactions~\cite{TGFuse,SHIP,LRRNet}, and disentangled representation learning~\cite{FreeFusion,DCDR-GAN}. However, most of these purely data-driven approaches rely on manually designed loss functions and heuristic architectural constraints to achieve fusion objectives~\cite{RXDNFuse,adversarial}. More comprehensive reviews and comparisons of IVIF methods can be found in~\cite{Data-Compatibility}.

\subsection{Prior Knowledge Integration in Image Fusion}
Prior-guided fusion has emerged as a pivotal paradigm in IVIF, substantially advancing both perceptual quality and downstream task performance~\cite{SDSFusion, PromptFusion, text-if,DCEvo}. A dominant trend exploits high-level semantics from large-scale foundation models. Representative works include SAGE~\cite{SAGE}, which harnesses Segment Anything Model (SAM)~\cite{SegmentAnything} priors via semantic-persistent attention and dual-level distillation, and CMFS~\cite{CMFS,SPDFusion}, which employs CLIP~\cite{clip} guidance for cross-modal interaction and frequency-spatial collaboration to suppress noise.

Complementary to foundation models, another line of research integrates priors from specialized vision tasks. STDFusionNet~\cite{STDFusionNet}, for example, leverages saliency detection masks for spatial guidance and selective feature extraction. Others utilize semantic guidance from pre-trained segmentation models~\cite{SPDFusion,SuperFusion,SeaFusion,segmif} or employ object detection meta-features to mitigate feature misalignment~\cite{MetaFusion,tardal}. Beyond task-level semantics, alternative prior formulations are embedded within the network architecture. These include incorporating low-rank representation priors to guide feature decomposition~\cite{LRRNet}, exploiting visible image quality priors~\cite{PIAFusion}, and exploring modality-specific priors within hybrid CNN-Transformer frameworks~\cite{Explainable}.

In summary, existing prior-guided IVIF methods often rely on rigid or extrinsic priors, which may cause optimization conflicts or granularity mismatch, leaving room for more adaptive, modality-intrinsic prior modeling.

\begin{figure*} 
  \centering
  \includegraphics[width=\linewidth]{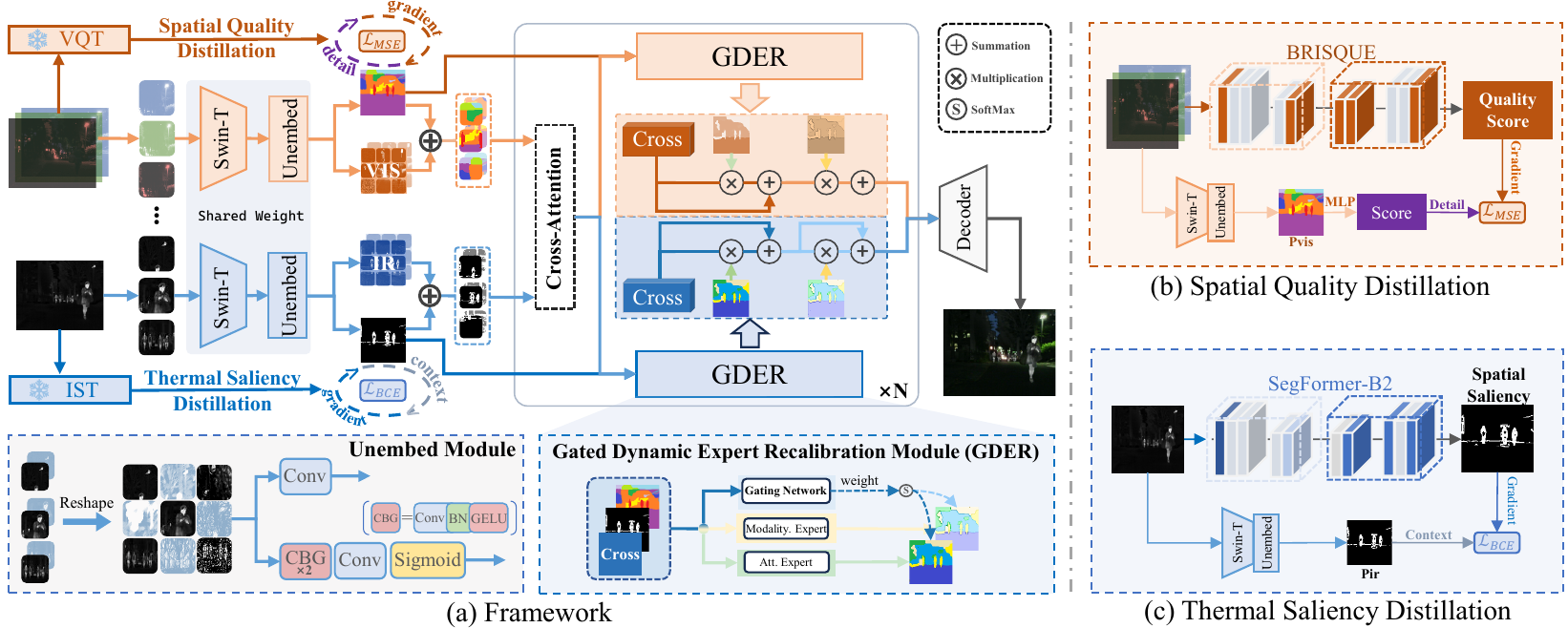} 
  \caption{\textbf{Illustration of the proposed framework.} (a) Overall pipeline: prior-guided visible and infrared features are fused via cross-attention, where the proposed GDER module adaptively refines modality-specific and interaction features for reconstruction. (b) Spatial Quality Distillation: a quality assessor provides global quality guidance to supervise spatial-detail learning. (c) Thermal Saliency Distillation: a saliency teacher offers thermal saliency priors to enhance target-aware fusion.}
  \label{fig:main}
\end{figure*}

\section{Methodology}
\label{methods}
In this section, we elucidate the proposed \text{P\textsuperscript{2}Fusion} framework, which instantiates the "Teach-to-Fuse" paradigm. The architecture, as illustrated in Fig.~\ref{fig:main}, consists of two core stages: (1) \textbf{Dual-Teacher Prompt Distillation}, which transforms static priors into learnable, process-oriented guidance signals; and (2) \textbf{Adaptive Fusion with Gated Dynamic Expert Recalibration (GDER)}, which achieves robust feature integration and self-correcting refinement via decoupled specialized experts.

\subsection{Dual-Teacher Prompt Distillation}
\label{subsec:Dual-Teacher}
In conventional prior-guided methods, knowledge (e.g., saliency maps) is typically injected as rigid spatial constraints, which limits the optimization flexibility of the network. We argue that prior knowledge should participate in the fusion process as soft, dynamic prompts. We select two complementary intrinsic priors as a validation instance:
\begin{itemize}
    \item Infrared Saliency Prior ($T_{ir}$): Derived from a SegFormer-B2~\cite{segformer} pre-trained on the representative MSRS~\cite{MSRS} and FMB~\cite{segmif} datasets, this prior serves as the Infrared-Saliency Teacher (IST) to localize high-thermal-radiation targets (e.g., pedestrians and vehicles) in the infrared modality.
    \item Visible Quality Prior ($T_{vis}$): Generated by the no-reference Visible-Quality Teacher (VQT) based on the BRISQUE~\cite{BRISQUE} evaluator, this prior assesses textural reliability spatially, such as underexposed or overexposed areas.
\end{itemize}

For the visible branch, the VQT generates a global quality score $T_{vis} \in \mathbb{R}$. We project the high-dimensional prompt $P_{vis}$ into a scalar $P'_{vis}$ to match the teacher's output. For the infrared branch, the IST provides a saliency map $T_{ir} \in \mathbb{R}^{H \times W}$. The final enriched features $F = \{F_{vis}, F_{ir}\}$ are obtained via element-wise summation: $F_m = f_m + \text{Embed}(P_m)$, where $m \in \{vis, ir\}$.

\begin{figure*} 
  \centering
  \includegraphics[width=\linewidth]{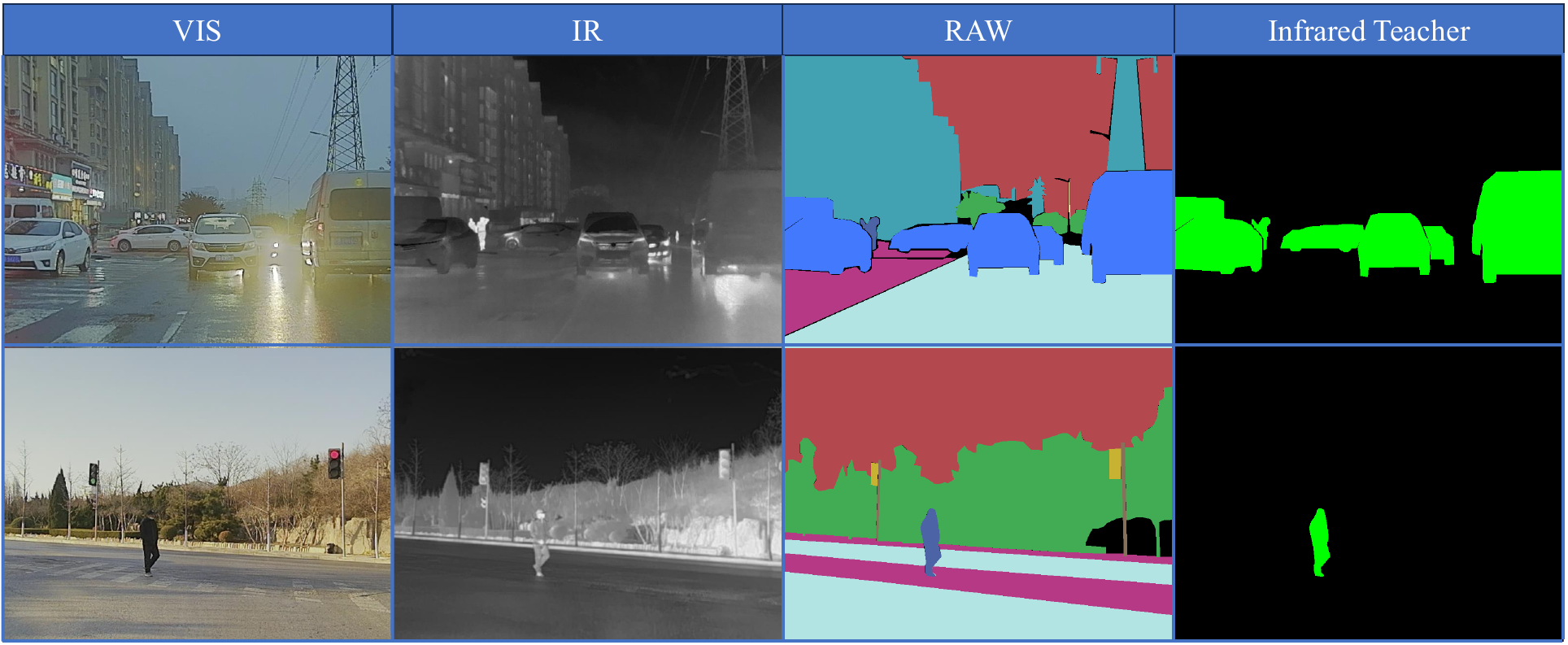} 
  \caption{Illustration of the difference between the original dataset and that used for training our infrared thermal saliency teacher: pedestrians and vehicles are defined as foreground, while all others are treated as background.}
  \label{fig:irT}
\end{figure*}

Specifically, for the infrared branch,semantic predictions are post-processed into binary saliency masks  to incorporate domain priors,  where pedestrians and vehicles are treated as foreground and all remaining categories as background as shown in Fig.~\ref{fig:irT}. The resulting saliency map is then downsampled via bilinear interpolation to match the spatial resolution of the infrared prompt $P_{ir}$. For the visible branch, given an image-level score $s \in [0,100]$, we normalize it into a continuous quality prior by $T_{vis} = (100 - s)/100$, where higher values correspond to stronger textural reliability.

\subsection{Adaptive Fusion with GDER Refinement}

The feature spaces of infrared and visible modalities are inherently heterogeneous—one being sensitive to thermal contours and the other rich in textural details. To mitigate modality confusion, we maintain dual-branch processing and employ cross-attention (CA)~\cite{transformer,swintransformer} mechanism for feature modulation:
\begin{equation}
F'_{ir} = \text{CA}(F_{ir}, F_{vis}, F_{vis}), \quad F'_{vis} = \text{CA}(F_{vis}, F_{ir}, F_{ir})
\end{equation}
where $\text{CA}(Q,K,V) = \text{softmax}\left(\frac{QK^\top}{\sqrt{d}}\right)V$. This stage yields cross-modal interactions $F' = \{F'_{vis}, F'_{ir}\}$. To further refine the fused representations and capture modal-specific nuances, we propose the Gated Dynamic Expert Recalibration (GDER) module. Unlike  Mixture-of-Experts (MoE) architectures designed primarily for model capacity expansion, our GDER module functions as a decoupled feature recalibrator with explicit functional specialization, aiming to adaptively aggregate the most informative cues from heterogeneous feature spaces:

\begin{enumerate}
    \item \textbf{Prior-Responsive Modality Expert ($E_{mod}$):} This expert explicitly consumes the dynamic prompts to reinforce modality-specific salient information (e.g., thermal targets or high-contrast textures). It ensures that the network maintains a high-fidelity focus on regions indicated by the learned priors, facilitating effective knowledge injection.
    \item \textbf{Prior-Agnostic Attention Expert ($E_{att}$):} Operating independently of the prompts, this expert utilizes a global attention mechanism (CBAM) to mine intrinsic long-range dependencies and fine-grained structural details. It acts as a contextual compensator to ensure the integrity of textures that may fall outside the prior's primary focus.
\end{enumerate}

The GDER module acts as an adaptive integration arbiter. A gating network $G(\cdot)$ computes dynamic weights $w = \text{softmax}(G(F', P))$ based on the mutual agreement between local feature characteristics and global prompt signals. The refined feature is formulated as:
\begin{equation}
F'' = F' + w_1 \cdot E_{mod}(F') + w_2 \cdot E_{att}(F')
\end{equation}
By synergizing the outputs of these decoupled specialists, the GDER module adaptively recalibrates bimodal information flow. The Prior-Responsive Expert anchors features on saliency-guided regions, while the Prior-Agnostic Expert captures the underlying structural dependencies that might be overlooked by prompts. This dual-expert design allows the network to transcend the limitations of any single representation, achieving robust feature recalibration that balances salient foreground targets with rich background textures.

In this configuration, the gating network acts as a dynamic weight allocator that optimizes the integration of modality-specific and context-aware features. When both specialists provide congruent information, they work in synergy to enhance feature representational power; when their focuses diverge, the gating mechanism adaptively balances their contributions to ensure that the fused feature remains both semantically meaningful and structurally complete. This mechanism achieves adaptive feature refinement through multi-perspective complementarity, rather than rigid constraint. After $N$ iterative refinements, the refined dual-modality features are concatenated and fed into the reconstruction decoder to generate the final fused image $I_{fused}$.

\subsection{Optimization Objectives}

We replace traditional hard-coded prior losses with learnable prompt guidance. The total loss $\mathcal{L}_{total}$ comprises perceptual constraints and distillation terms.

\textbf{Perceptual Constraints:} The Gradient Loss ensures edge preservation: $\mathcal{L}_{grad} = \| \nabla I_{fused} - \max(|\nabla I_{ir}|, |\nabla I_{vis}|) \|_1$. The Intensity Loss balances thermal and texture information: $\mathcal{L}_{int} = \| I_{fused} - \max(I_{ir}, I_{vis}) \|_1$. We further introduce the SSIM Loss to maintain structural consistency:
\begin{equation}
\mathcal{L}_{ssim} = 1 - (\omega_{ir} \text{SSIM}(I_{ir}, I_{fused}) + \omega_{vis} \text{SSIM}(I_{vis}, I_{fused}))
\end{equation}
where weights $\omega$ are adaptively computed based on the mean gradient magnitude of each modality.

\textbf{Prompt Distillation Loss:} The visible prompts are optimized via MSE: $\mathcal{L}_{vis}^{distill} = \|P'_{vis} - T_{vis}\|_2^2$. The infrared prompts are optimized via Binary Cross-Entropy (BCE) to ensure spatial saliency alignment:
\begin{equation}
\mathcal{L}_{ir}^{distill} = -(T_{ir} \odot \log P_{ir}^{map} + (1 - T_{ir}) \odot \log(1 - P_{ir}^{map}))
\end{equation}
The overall optimization objective is defined as:
\begin{equation}
\mathcal{L}_{total} = \lambda_1 \mathcal{L}_{int} + \lambda_2 \mathcal{L}_{ssim} + \lambda_3 \mathcal{L}_{grad} + \lambda_4 \mathcal{L}_{ir}^{distill} + \lambda_5 \mathcal{L}_{vis}^{distill}
\end{equation}

\section{Experiments}
\label{sec:Experiments}
\subsection{Experiments Settings}
\label{subsec:Experiments Settings}
\noindent\textbf{Datasets and metrics.} We evaluate P\textsuperscript{2}Fusion on five public benchmarks: MSRS~\cite{MSRS}, M3FD~\cite{tardal}, FMB~\cite{segmif}, RoadScene~\cite{U2Fusion} and DroneVehicle~\cite{drone}. Generalization is assessed on their official test splits, covering fusion quality (all), object detection (MSRS, M3FD, DroneVehicle), and semantic segmentation (MSRS, FMB). Fusion quality is measured using four standard metrics, while downstream evaluations follow the respective standard protocols.

\noindent\textbf{Implementations.}  Our framework is implemented in PyTorch on four NVIDIA RTX 3090 GPUs. Adam optimizes parameters for 100 epochs (batch size 12, initial lr $1 \times 10^{-4}$), decayed by 0.5 at iterations [3000, 6000, 9000, 12000, 15000]. The optimal weights for the total loss ($\lambda$1–$\lambda$5) are (8,25,20,0.5,5).

\noindent\textbf{Comparisons.} We compare with 12 SOTA fusion networks (SAGE~\cite{SAGE}, DitFuse~\cite{ditfuse2025}, MRFS~\cite{MRFS}, Freefusion~\cite{FreeFusion}, FreqGAN~\cite{FreqGAN}, LutFuse~\cite{LUT-Fuse}, LRRNet~\cite{LRRNet}, DDFM~\cite{DDFM}, TarDal~\cite{tardal}, ReCoNet~\cite{ReCoNet}, TIM~\cite{TIM}) to validate our method’s effectiveness and generalization ability.

\begin{figure}[!t]
  \centering
  \includegraphics[width=\linewidth]{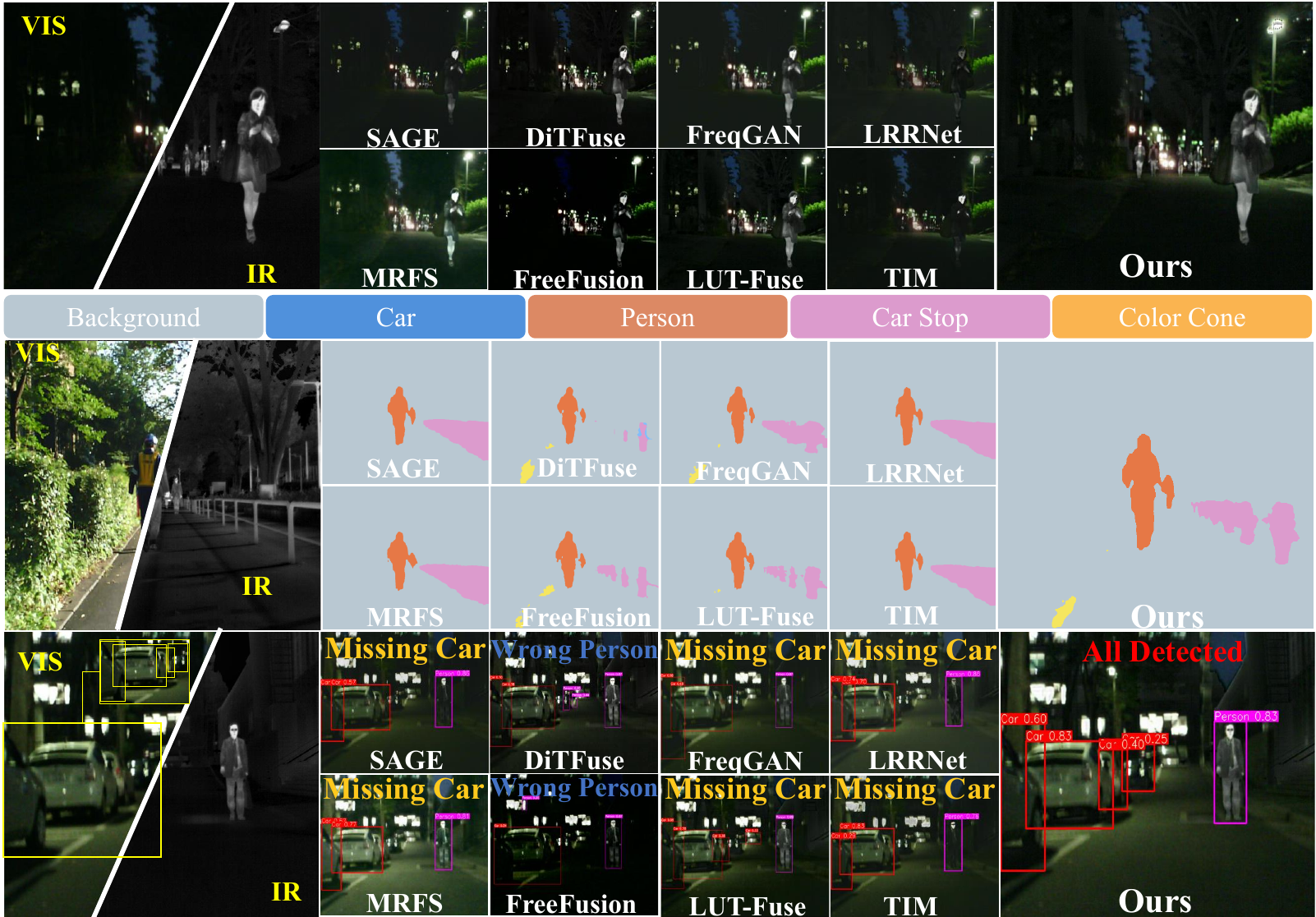} 
  \caption{\textbf{Comprehensive qualitative comparison on the MSRS dataset.} Row 1 shows fused images; Row 2 and Row 3 show the corresponding semantic segmentation and object detection results, respectively.  Our method consistently achieves superior performance across image fusion, semantic segmentation, and object detection tasks.}
  \label{fig:fig2}
\end{figure}

\begin{figure}[!t]
  \centering
  \includegraphics[width=\linewidth]{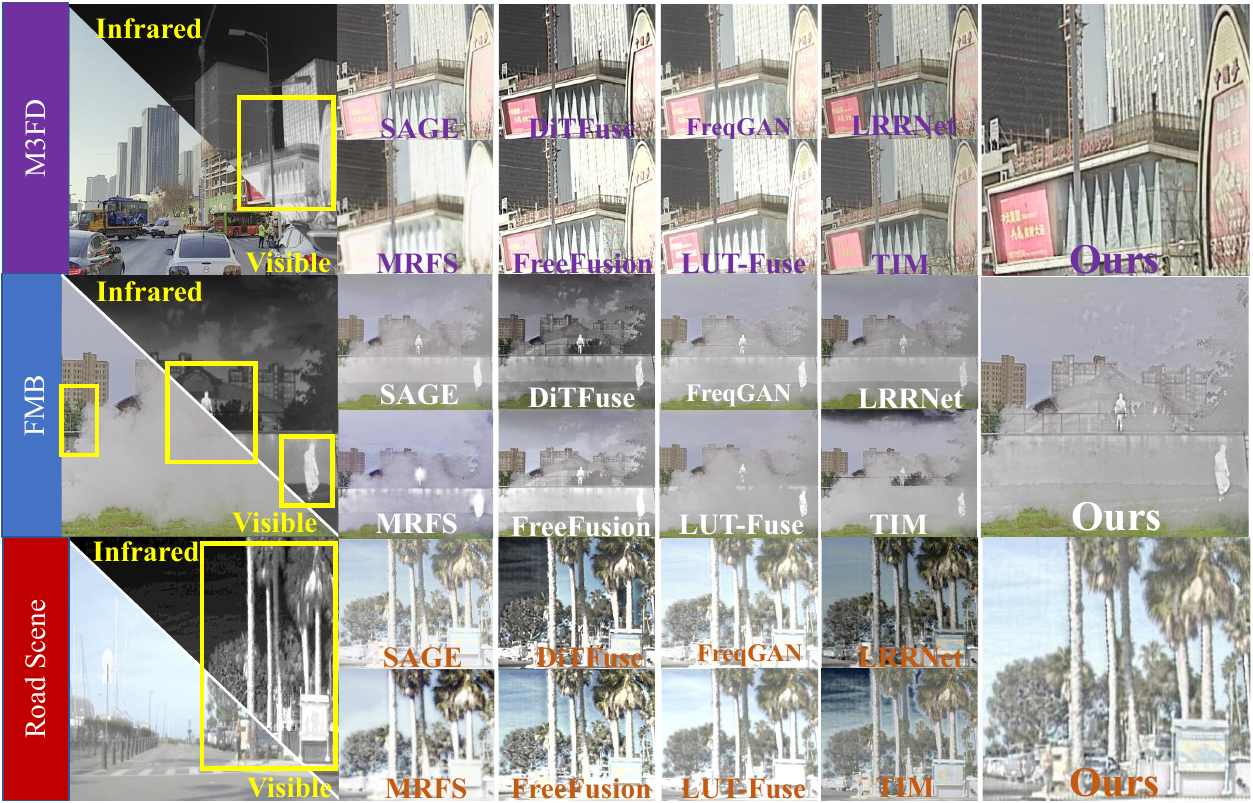} 
  \caption{\textbf{Qualitative fusion comparison.} Representative high-contrast scenes from M3FD, smoke-occluded scenes from FMB, and over-exposed scenes from RoadScene are selected. Our method consistently preserves infrared salient targets and visible textures across degradations, achieving the best modality balance.}

  \label{fig:fig3}
\end{figure}

\subsection{Infrared and visible image fusion}
\label{Infrared and visible image fusion}
We conduct IVIF experiments on MSRS (361 pairs), M3FD (300), FMB (280), RoadScene (221) and DroneVehicle (8980). Training uses combined MSRS (1083 pairs) and FMB (1220) training sets, whose segmentation labels train the infrared saliency teacher as mentioned in ~\ref{subsec:Dual-Teacher}.

\textbf{Qualitative Comparisons.} Qualitative results for selected pairs are in Fig.~\ref{fig:fig2} (first row) and Fig.~\ref{fig:fig3}. On the MSRS dataset, SAGE, DiTFuse, FreqGAN, LRRNet, FreeFusion, and TIM exhibit an over-reliance on visible textures at the expense of infrared saliency, resulting in attenuated structural details in low-light regions. Conversely, MRFS yields over-smoothed textures with suboptimal local contrast, while LUT-Fuse introduces noticeable artifacts around moving targets despite better bimodal preservation. In contrast, our method effectively recovers background details (e.g., tree branches and shadowed building textures) while maintaining sharp pedestrian contours and high fidelity.

In the high-contrast M3FD scene (Fig.~\ref{fig:fig3}, first row), competing methods suffer from modality collapse, making the background text indistinct. Our method maintains a stable equilibrium, yielding clearer structures and superior text legibility. 

In the smoke-occluded FMB scene (Fig.~\ref{fig:fig3}, second row), most competitors (SAGE, MRFS, FreqGAN, LUT-Fuse, LRRNet, TIM) under-exploit infrared cues, leaving details corrupted by smoke. Meanwhile, DiTFuse and FreeFusion over-rely on the infrared modality, inducing an unnatural night-like shift and losing visible textures like the left tree. Our method ensures an optimal balance, simultaneously preserving infrared contours and fine visible details.

In the over-exposed RoadScene scene (Fig.~\ref{fig:fig3}, third row), SAGE, MRFS, FreeFusion, FreqGAN, and LUT-Fuse under-utilize infrared cues, resulting in over-exposed outputs that miss critical contours (e.g., the pedestrian near the tree trunk). In contrast, DiTFuse, LRRNet, and TIM fail to maintain modality balance, producing unnaturally dark skies and losing details on billboards. Our method achieves a more stable dynamic balance, recovering the global appearance while preserving rich textures and fine details.

\textbf{Quantitative Comparisons.} Tables \ref{tab:1} report the quantitative results. Across all datasets, our method achieves SOTA on 12 out of 16 metrics and ranks second on 3 others, highlighting its overall superiority and strong cross-dataset generalization.

\begin{table*}[!t]
\centering
\caption{Quantitative comparison on the \textbf{MSRS}, \textbf{M3FD}, \textbf{FMB}, and \textbf{RoadScene} datasets using four key evaluation metrics. (1st: \textbf{\textcolor{top1color}{Red}}, 2nd: \textbf{\textcolor{top2color}{Light Blue}})}
\label{tab:1}
\scriptsize
\begin{adjustbox}{width=\textwidth}
\begin{tabular}{@{} l|cccc|cccc|cccc|cccc @{}}
\toprule
\textbf{Datasets} & \multicolumn{4}{c|}{\textbf{MSRS}} & \multicolumn{4}{c|}{\textbf{M3FD}} & \multicolumn{4}{c|}{\textbf{FMB}} & \multicolumn{4}{c}{\textbf{RoadScene}} \\
\midrule
\textbf{Methods} & $FQI_{E}$$\uparrow$ & $VIF$$\uparrow$ & $Q_{abf}$$\uparrow$ & $MI$$\uparrow$ & $FQI_{E}$$\uparrow$ & $VIF$$\uparrow$ & $Q_{abf}$$\uparrow$ & $MI$$\uparrow$ & $FQI_{E}$$\uparrow$ & $VIF$$\uparrow$ & $Q_{abf}$$\uparrow$ & $MI$$\uparrow$ & $FQI_{E}$$\uparrow$ & $VIF$$\uparrow$ & $Q_{abf}$$\uparrow$ & $MI$$\uparrow$ \\
\midrule
DiTFuse(TPAMI'25) & 0.642 & 0.281 & 0.363 & 2.146 & 0.389 & 0.188 & 0.274 & 2.514 & 0.536 & 0.215 & 0.363 & 2.620 & 0.445 & 0.320 & 0.426 & 2.938 \\
FreqGAN(TCSVT'25) & 0.679 & 0.285 & 0.492 & 2.871 & 0.445 & 0.282 & 0.399 & 3.272 & 0.518 & 0.293 & 0.437 & 3.434 & 0.442 & 0.279 & 0.444 & 2.967 \\
Freefusion(TPAMI'25) & 0.590 & 0.281 & 0.430 & 1.984 & 0.621 & \cellcolor{top1color}\textbf{0.473} & 0.539 & 2.712 & 0.696 & \cellcolor{top1color}\textbf{0.501} & 0.587 & 3.004 & 0.403 & 0.362 & 0.403 & 2.692 \\
LutFuse(ICCV'25) & \cellcolor{top2color}0.783 & \cellcolor{top2color}0.424 & \cellcolor{top2color}0.610 & \cellcolor{top2color}3.625 & 0.466 & 0.319 & 0.480 & \cellcolor{top1color}\textbf{3.987} & 0.575 & 0.319 & 0.527 & \cellcolor{top2color}3.897 & 0.328 & 0.265 & 0.381 & \cellcolor{top1color}\textbf{3.644} \\
SAGE(CVPR'25) & 0.763 & 0.335 & 0.535 & 3.217 & \cellcolor{top2color}0.659 & 0.363 & \cellcolor{top2color}0.575 & 3.115 & \cellcolor{top2color}0.763 & 0.384 & \cellcolor{top2color}0.634 & 3.429 & 0.379 & 0.237 & 0.334 & 2.919 \\
TIM(TPAMI'24) & 0.655 & 0.277 & 0.469 & 3.025 & 0.533 & 0.325 & 0.524 & 3.533 & 0.679 & 0.125 & 0.581 & 3.534 & 0.297 & 0.181 & 0.354 & \cellcolor{top2color}3.611 \\
MRFS(CVPR'24) & 0.625 & 0.340 & 0.484 & 3.068 & 0.287 & 0.194 & 0.249 & 2.965 & 0.477 & 0.161 & 0.388 & 3.087 & 0.268 & 0.159 & 0.271 & 2.788 \\
LRRNet(TPAMI'23) & 0.599 & 0.243 & 0.420 & 2.938 & 0.530 & 0.283 & 0.483 & 2.823 & 0.649 & 0.112 & 0.539 & 3.024 & 0.277 & 0.133 & 0.326 & 2.791 \\
DDFM(ICCV'23) & 0.569 & 0.363 & 0.469 & 2.666 & 0.522 & 0.311 & 0.456 & 2.851 & 0.566 & 0.214 & 0.504 & 3.144 & \cellcolor{top2color}0.485 & \cellcolor{top2color}0.372 & \cellcolor{top2color}0.471 & 2.940 \\

TarDal(CVPR'22) & 0.495 & 0.357 & 0.428 & 2.646 & 0.425 & 0.317 & 0.402 & 3.182 & 0.443 & 0.304 & 0.406 & 3.419 & 0.395 & 0.328 & 0.392 & 3.404 \\
ReCoNet(ECCV'22) & 0.720 & 0.325 & 0.500 & 3.115 & 0.570 & 0.291 & 0.481 & 3.095 & 0.692 & 0.168 & 0.546 & 3.288 & 0.422 & 0.245 & 0.334 & 3.139 \\
\midrule
\textbf{Ours} & \cellcolor{top1color}\textbf{0.850} & \cellcolor{top1color}\textbf{0.445} & \cellcolor{top1color}\textbf{0.682} & \cellcolor{top1color}\textbf{3.905} & \cellcolor{top1color}\textbf{0.743} & \cellcolor{top2color}0.429 & \cellcolor{top1color}\textbf{0.635} & \cellcolor{top2color}3.819 & \cellcolor{top1color}\textbf{0.826} & \cellcolor{top2color}0.448 & \cellcolor{top1color}\textbf{0.677} & \cellcolor{top1color}\textbf{4.030} & \cellcolor{top1color}\textbf{0.617} & \cellcolor{top1color}\textbf{0.388} & \cellcolor{top1color}\textbf{0.557} & 3.021 \\
\bottomrule
\end{tabular}
\end{adjustbox}
\end{table*}

\begin{figure}[!t]
  \centering
  \vspace{-2pt}
  \includegraphics[width=\linewidth]{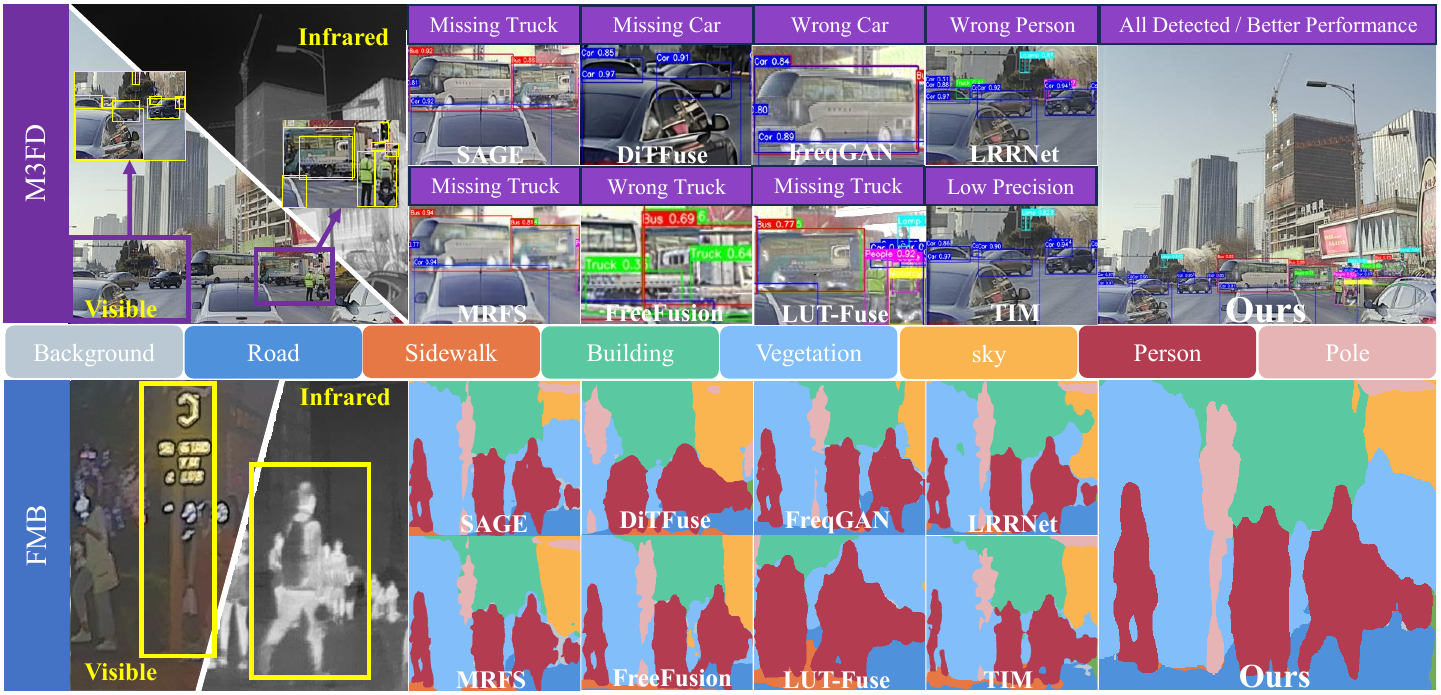} 
  \caption{\textbf{Qualitative evaluation on downstream tasks.} We visualize object detection results on M3FD (Row 1) and semantic segmentation results on FMB (Row 2). Our framework demonstrates superior performance in downstream tasks.}
  \vspace{-2pt}
  \label{fig:fig4}
\end{figure}

\begin{table*}[!t]
\centering
\caption{Quantitative results of downstream tasks: Object Detection on \textbf{M3FD} and Semantic Segmentation on \textbf{FMB}. (1st: \textbf{\textcolor{top1color}{Red}}, 2nd: \textbf{\textcolor{top2color}{Light Blue}})}
\label{tab:2}
\scriptsize
\begin{adjustbox}{width=\textwidth}
\begin{tabular}{@{} l|ccccccc|cccccccc @{}}
\toprule
\textbf{Datasets} & \multicolumn{7}{c|}{\textbf{M3FD(Object Detection)}} & \multicolumn{8}{c}{\textbf{FMB(Semantic Segmentation)}} \\ 
\midrule
\textbf{Methods} & Person & Car & Bus & Lamp & Moto. & Truck & \textbf{mAP}$\uparrow$ & T.Sign & Person & Car & Truck & Moto. & Pole & \textbf{mPA}$\uparrow$ & \textbf{mIoU}$\uparrow$ \\
\midrule
DiTFuse(TPAMI'25)    & 0.514 & 0.684 & 0.742 & 0.497 & 0.525 & 0.663 & 60.46 & 73.146 & 63.568 & 81.763 & 33.809 & \cellcolor{top1color}\textbf{38.908} & 43.214 & 63.340 & 56.808 \\
FreqGAN(TCSVT'25)   & \cellcolor{top1color}\textbf{0.544} & 0.700 & 0.766 & 0.510 & 0.519 & 0.673 & 61.85 & \cellcolor{top2color}73.558 & \cellcolor{top2color}66.301 & \cellcolor{top2color}82.225 & 31.691 & \cellcolor{top2color}38.425 & \cellcolor{top1color}\textbf{44.730} & \cellcolor{top2color}64.471 & \cellcolor{top2color}57.284 \\
Freefusion(TPAMI'25) & 0.527 & 0.702 & \cellcolor{top2color}0.770 & \cellcolor{top2color}0.526 & 0.512 & 0.674 &\cellcolor{top2color} 61.87 & 71.751 & 64.320 & 80.160 & \cellcolor{top2color}36.174 & 33.359 & 41.968 & 63.469 & 56.421 \\
LutFuse(ICCV'25)     & 0.530 & 0.691 & 0.763 & 0.500 & 0.501 & 0.670 & 60.91 & 72.434 & 65.192 & 80.893 & 16.908 & 38.130 & 42.354 & 62.221 & 55.113 \\
SAGE(CVPR'25)        & 0.531 & \cellcolor{top1color}\textbf{0.707} & 0.735 & 0.483 & \cellcolor{top2color}0.539 & 0.658 & 60.86 & 72.390 & 65.286 & 81.387 & 31.706 & 35.223 & 41.864 & 63.798 & 56.479 \\
TIM(TPAMI'24)        & 0.496 & 0.686 & 0.743 & 0.503 & 0.506 & 0.665 & 59.97 & 72.216 & 62.701 & 81.081 & 26.209 & 30.750 & 41.752 & 61.942& 54.776 \\
MRFS(CVPR'24)        & 0.508 & 0.687 & 0.760 & 0.482 & 0.483 & 0.657 & 59.68 & 67.617 & 63.632 & 79.955 & 28.056 & 33.916 & 40.510 & 60.926& 53.959 \\
LRRNet(TPAMI'23)     & 0.519 & 0.693 & 0.756 & 0.514 & 0.499 & 0.659 & 60.67 & 70.693 & 64.877 & 81.654 & 34.215 & 31.684 & 41.634 & 63.853& 56.382 \\
DDFM(ICCV'23)        & 0.526 & \cellcolor{top2color}0.706 & 0.749 & 0.483 & \cellcolor{top1color}\textbf{0.558} & 0.662 & 61.44 & 71.478 & 65.847 & 81.797 & 32.143 & 33.463 & 42.865 & 64.466& 57.001 \\

TarDal(CVPR'22)      & \cellcolor{top1color}\textbf{0.544} & 0.693 & 0.755 & 0.512 & 0.495 & 0.672 & 61.22 & 72.249 & \cellcolor{top1color}\textbf{66.804} & 80.383 & 15.760 & 34.881 & 41.201 & 62.462& 55.193 \\
ReCoNet(ECCV'22)     & 0.525 & 0.704 & 0.755 & \cellcolor{top1color}\textbf{0.529} & 0.500 & \cellcolor{top2color}0.679 & 61.53 & 71.570 & 64.709 & 81.360 & 28.587 & 32.394 & 41.458 & 63.631& 56.113 \\
\midrule
\textbf{Ours}        & \cellcolor{top2color}0.534 & 0.704 & \cellcolor{top1color}\textbf{0.773} & 0.521 & 0.521 & \cellcolor{top1color}\textbf{0.690} & \cellcolor{top1color}\textbf{62.37} &\cellcolor{top1color}\textbf{74.153} & 65.670 & \cellcolor{top1color}\textbf{82.270} & \cellcolor{top1color}\textbf{38.270} & 37.024 & \cellcolor{top2color}43.913& \cellcolor{top1color}\textbf{64.672} & \cellcolor{top1color}\textbf{57.456} \\
\bottomrule
\end{tabular}
\end{adjustbox}
\end{table*}

\begin{table*}[!t]
\centering
\caption{Quantitative comparison on the MSRS datasets for object detection and segmentation. (1st: \textbf{\textcolor{top1color}{Red}}, 2nd: \textbf{\textcolor{top2color}{Light Blue}})}
\label{tab:3}
\small
\begin{adjustbox}{width=\textwidth}
\begin{tabular}{@{} l|cc|cc|cc|ccccccc @{}}
\toprule
\textbf{Tasks} & \multicolumn{6}{c|}{\textbf{Object Detection ($mAP_{50} \mid mAP_{50\to95}$)}} & \multicolumn{7}{c}{\textbf{Semantic Segmentation}} \\
\midrule
\textbf{Methods} & \multicolumn{2}{c|}{\textbf{Person}} & \multicolumn{2}{c|}{\textbf{Car}} & \multicolumn{2}{c|}{\textbf{All}} & \textbf{Car} & \textbf{Person} & \textbf{Curve} & \textbf{Stop} & \textbf{Cone} & \textbf{mPA}$\uparrow$ & \textbf{mIoU}$\uparrow$ \\
\midrule
DiTFuse(TPAMI'25)    & 0.993 & 0.717 & 0.838 & \cellcolor{top2color}0.715 & 0.915 & \cellcolor{top2color}0.716 & 88.073 & 68.929 & \cellcolor{top2color}40.988 & 53.430 & \cellcolor{top2color}66.992 & 70.555 & 62.390\\
FreqGAN(TCSVT'25)   & \cellcolor{top1color}\textbf{0.995} & 0.731 & 0.906 & 0.645 & 0.951 & 0.688 &87.681 & 69.427 & 33.235 & \cellcolor{top1color}\textbf{62.014} & 66.369 & 70.149 & 62.100 \\
Freefusion(TPAMI'25) & 0.991 & 0.723 & 0.763 & 0.385 & 0.877 & 0.554 &  84.702 & 66.407 & 36.520 & 64.899 & 64.334 & 68.776 & 61.120  \\
LutFuse(ICCV'25)     & \cellcolor{top1color}\textbf{0.995} & 0.727 & 0.937 & 0.628 & 0.966 & 0.677 &87.848 & \cellcolor{top1color}\textbf{70.602} & \cellcolor{top1color}\textbf{44.679} & 59.176 & 67.835 & \cellcolor{top1color}\textbf{71.558} & \cellcolor{top2color}63.471\\
SAGE(CVPR'25)        & 0.991 & 0.700 & 0.609 & 0.476 & 0.800 & 0.588 & 91.265 & 66.107 & 27.288 & 29.242 & 56.066 & 63.055 & 56.228 \\
TIM(TPAMI'24)        & 0.986 & 0.603 & 0.920 & 0.643 & 0.953 & 0.623 & 89.800 & 62.801 & 31.877 & 32.128 & 56.872 & 61.471 & 55.267 \\
MRFS(CVPR'24)        & 0.991 & 0.738 & 0.920 & 0.614 & 0.955 & 0.676 & 91.222 & 65.075 & 26.845 & 29.521 & 56.422 & 62.333 & 55.073 \\
LRRNet(TPAMI'23)     & \cellcolor{top2color}0.994 & \cellcolor{top2color}0.757 & \cellcolor{top2color}0.962 & 0.606 & \cellcolor{top2color}0.978 & 0.681 & 91.705 & 67.741 & 30.996 & 31.368 & 57.681 & 62.450 & 56.567 \\
DDFM(ICCV'23)        & 0.986 & 0.697 & 0.747 & 0.585 & 0.866 & 0.641 & \cellcolor{top1color}\textbf{91.911} & 68.686 & 31.325 & 28.690 & 56.075 & 63.932 & 56.875 \\
TarDal(CVPR'22)      & 0.974 & 0.716 & 0.858 & 0.669 & 0.916 & 0.692 & 91.304 & 68.907 & 28.039 & 28.831 & 55.925 & 62.784 & 56.534 \\
ReCoNet(ECCV'22)     & \cellcolor{top2color}0.994 & 0.712 & 0.866 & 0.573 & 0.931 & 0.643 & \cellcolor{top2color}91.750 & 66.247 & 30.113 & 32.249 & 56.779 & 62.844 & 56.928 \\
VIS                  & 0.851 & 0.477 & 0.862 & 0.626 & 0.856 & 0.552 & 91.667 & 57.770 & 25.817 & 29.342 & 59.742 & 60.604 & 54.559 \\
IR                   & 0.979 & 0.613 & \cellcolor{top2color}0.962 & 0.672 & 0.971 & 0.643 & 87.155 & 68.749 & 25.281 & 22.444 & 45.700 & 54.172 & 48.778 \\
\midrule
\textbf{Ours}        & \cellcolor{top1color}\textbf{0.995} & \cellcolor{top1color}\textbf{0.776} & \cellcolor{top1color}\textbf{0.995} & \cellcolor{top1color}\textbf{0.720} & \cellcolor{top1color}\textbf{0.995} & \cellcolor{top1color}\textbf{0.748} & 88.520 & \cellcolor{top2color}70.030 & 40.903& \cellcolor{top2color}59.800 & \cellcolor{top1color}\textbf{69.494} & \cellcolor{top2color}71.480 & \cellcolor{top1color}\textbf{63.793}  \\
\bottomrule
\end{tabular}
\end{adjustbox}
\end{table*}

\subsection{IVIF for Downstream Tasks}
\label{IVIF for Downstream Tasks}
For object detection, we train YOLOv7s~\cite{YOLOv7} on MSRS-detection (80 pairs) and M3FD (4200 pairs) with an 8:2 split. Tables~\ref{tab:2} (left) and~\ref{tab:3} (left) show that our method achieves the highest mAP$_{50\text{--}95}$, outperforming the previous SOTA by \textbf{+3.2\%} on MSRS and \textbf{+0.5\%} on M3FD. Visualizations in Figs.~\ref{fig:fig2} (third row) and~\ref{fig:fig4} confirm that our fused images successfully detect all scene targets, whereas other methods suffer from false positives or missed detections.

For semantic segmentation, SegFormer-B5~\cite{segformer} is adopted on MSRS (1083/361 pairs) and FMB (1220/280). As reported in Tables~\ref{tab:2} (right) and~\ref{tab:3} (right), our framework delivers the best performance with consistent gains. Qualitative results (Figs.~\ref{fig:fig2}, \ref{fig:fig4}, second rows) demonstrate more coherent masks with better structural preservation and reduced fragmentation.

\subsection{Ablation Study}
\label{Ablation Study}
\begin{figure}[!t]   
    \includegraphics[width=\linewidth]{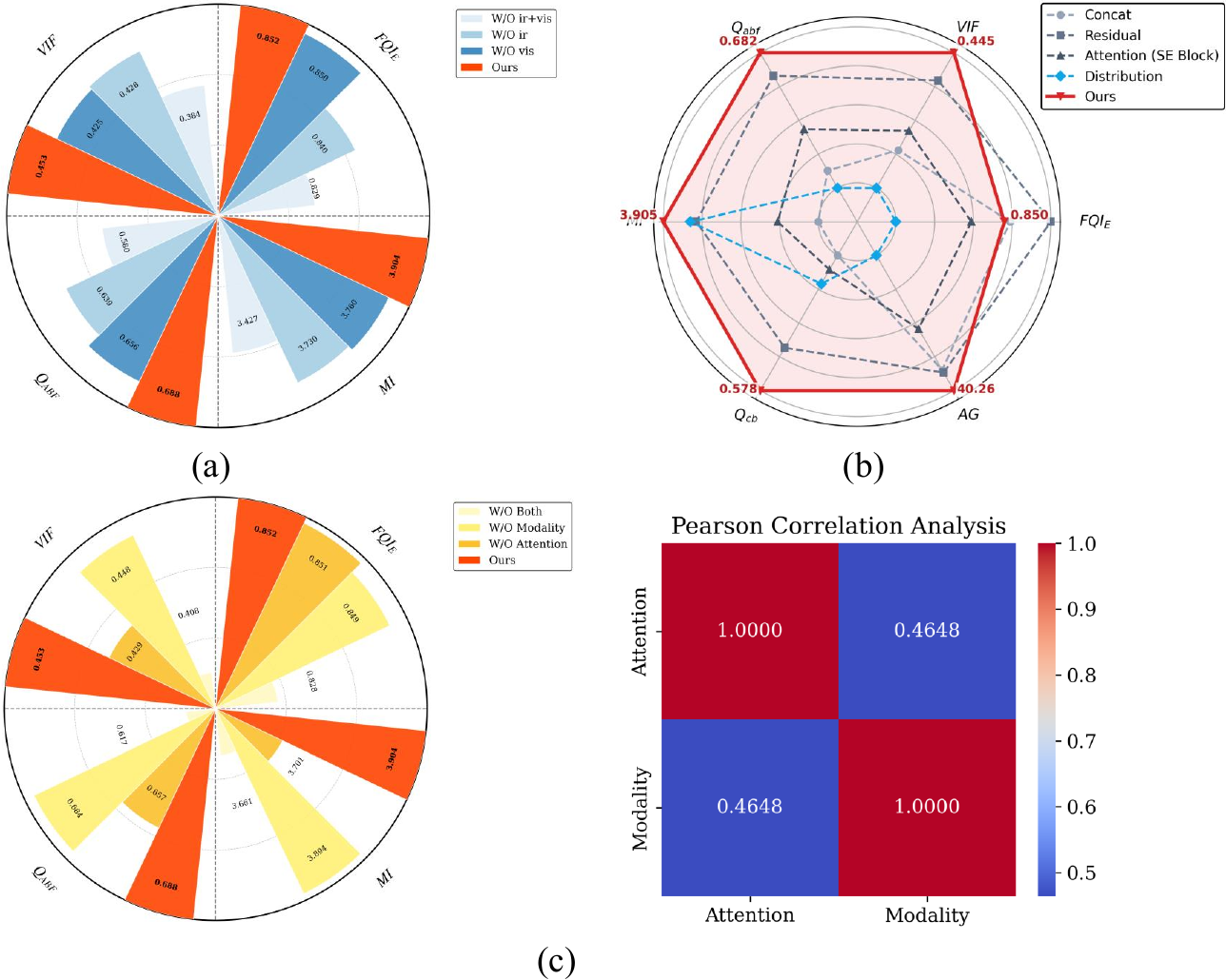}  
    \caption{\textbf{Ablation study of the proposed components.} All metrics are normalized relative to the full model (Ours). 
(a) \textbf{Impact of image priors:} Dual priors show clear synergy in maintaining visual fidelity. 
(b) \textbf{Structural alternatives to GDER:} Comparison with simpler architectural designs. 
(c) \textbf{Analysis of GDER teachers:} (Left) Performance gains from Modality and Attention teachers. (Right) Correlation analysis verifying the functional decoupling and non-redundancy between experts.}
    \label{fig:fig5}
\end{figure}

\noindent\textbf{Study on Prior Knowledge Integration.}
Fig.~\ref{fig:fig5}(a) further demonstrates that removing either prior consistently degrades performance across all metrics. Specifically, excluding the visible prior (W/O vis) causes a sharp drop in $VIF$, highlighting its importance in preserving fine-grained textures and structural fidelity. In contrast, the full ablation of both infrared and visible priors (W/O ir+vis) yields the worst performance, indicating that prior guidance is essential for balanced fusion quality.

\begin{figure}[!t]  
    \centering
    \includegraphics[width=\linewidth]{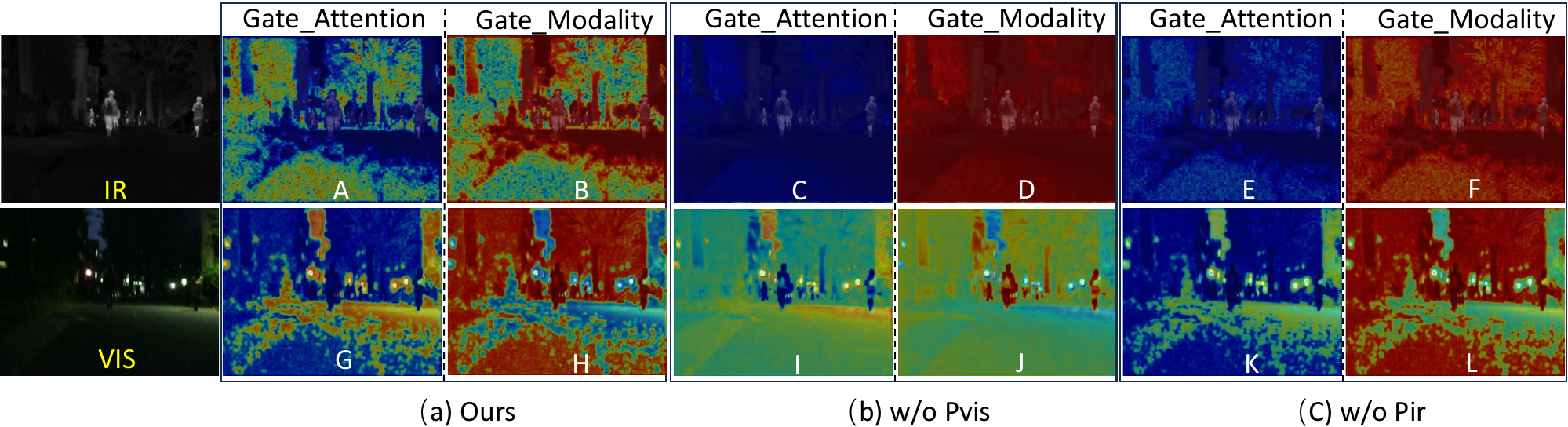}  
    \caption{\textbf{Visual analysis of the gating mechanism.} Visualization of cross-modal gating allocations, where ablating $P_{vis}$ or $P_{ir}$ disrupts spatial discrimination and leads to uniform or blurred feature distributions.}
    \label{fig:fig1-2}  
    \vspace{-10pt}
\end{figure}

Fig.~\ref{fig:fig1-2} demonstrates the effectiveness of bimodal priors in orthogonal gating. The full model (a) achieves structured, spatially selective gating. Removing $P_{vis}$ (b) causes the gating to collapse into a near-uniform distribution, confirming its role in providing localized guidance. Conversely, removing $P_{ir}$ (c) results in over-smoothed IR responses and a loss of cross-modal enhancement, highlighting its importance for discriminative performance.

\noindent\textbf{Study on Gated Dynamic Expert Recalibration.}
Fig.~\ref{fig:fig5}(b) compares our GDER module against intuitive alternatives, such as applying patch-level rather than global-level loss to $P_{\text{vis}}$ ("Distribution"). Empirical results confirm the superiority of our approach. Fig.~\ref{fig:fig5}(c) validates the GDER teacher framework, where the "W/O both" configuration shows the largest performance drop, underscoring the architecture's importance. Additionally, a Pearson correlation of $r=0.46$ between the Modality and Attention experts demonstrates effective functional decoupling. Our full model achieves the optimal score across all metrics, confirming that our dual-teacher strategy provides comprehensive, non-redundant guidance.

\begin{figure}[!t]  
    \centering
    \includegraphics[width=\linewidth]{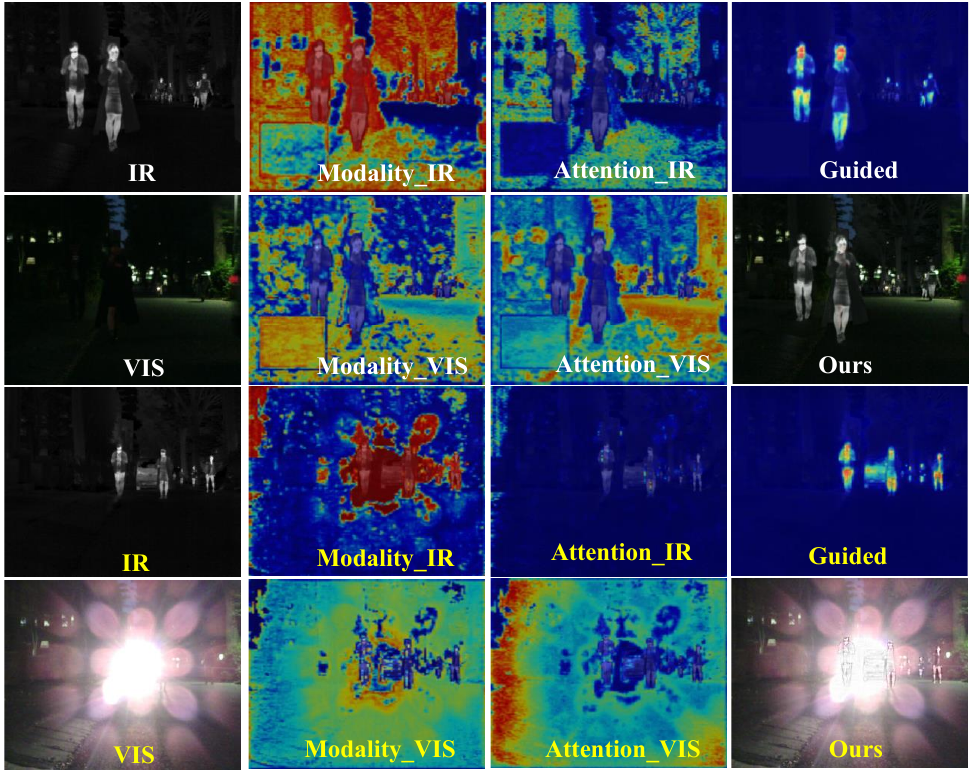}  
    \caption{\textbf{Qualitative analysis of expert decoupling and feature visualization.} We present the activation maps for the Modality and Attention experts across infrared and visible branches. The distinct focus of each expert—targeting either thermal saliency or structural textures—validates the functional complementarity of our MoE-based decoupling strategy.}
    
    \label{fig:fig6}  
    \vspace{-10pt}
\end{figure}

Fig.~\ref{fig:fig6} visualizes expert activation maps and prior-guided prompts. Complementarity within one modality and distinct responses across modalities validate our decoupled specialization over naive fusion. In overexposed scenes, the attention expert captures structural details while the modality expert isolates thermal targets, proving they are synergistically coupled yet representationally decoupled. Guided by high-fidelity spatial anchors, our framework ensures robust cross-modal integration.

\begin{table*}[!t]
\centering
\caption{Quantitative results of OOD Verification on the DroneVehicle, including image fusion and downstream object detection results. (1st: \textbf{\textcolor{top1color}{Red}}, 2nd: \textbf{\textcolor{top2color}{Light Blue}})}
\label{tab:ood}
\scriptsize
\begin{adjustbox}{width=\textwidth}
\begin{tabular}{@{} l|cccc|ccccc @{}}
\toprule
\textbf{Tasks} & \multicolumn{4}{c|}{\textbf{Image Fusion Metrics}} & \multicolumn{5}{c}{\textbf{Object Detection}} \\
\cmidrule(r){1-1} \cmidrule(lr){2-5} \cmidrule(l){6-10}
\textbf{Methods} & $FQI_{E}$$\uparrow$ & $VIF$$\uparrow$ & $Q_{abf}$$\uparrow$ & $MI$$\uparrow$ & Car & Truck & Bus & Van & \textbf{ $mAP_{50\to95}$}$\uparrow$ \\
\midrule
DiTFuse (TPAMI'25)   & 0.3929 & 0.1556 & 0.3134 & 2.1892 & 0.382 & 0.354& 0.578 & 0.234 & 0.387 \\
FreqGAN (TCSVT'25)   & \cellcolor{top2color}0.5340 & \cellcolor{top2color}0.2775 & 0.4643 & 2.6642 & 0.601 & 0.490 & 0.710 & 0.360 & 0.540 \\
FreeFusion (TPAMI'25)& 0.4893 & 0.2690 & 0.4339 & 2.4806 & 0.558 & 0.377 & 0.659 & 0.260 & 0.463 \\
LutFuse (ICCV'25)    & 0.5203 & \cellcolor{top1color}\textbf{0.2955} & \cellcolor{top2color}0.4984 & \cellcolor{top1color}\textbf{3.8038} & 0.571& 0.411 & 0.681 & 0.292 & 0.489 \\
SAGE (CVPR'25)       & 0.4443 & 0.1791 & 0.3636 & 2.4162 & 0.582 & 0.447 & 0.694 & 0.324 & 0.512 \\
TIM (TPAMI'24)       & 0.2728 & 0.2211 & 0.3182 & 2.6823 & \cellcolor{top2color}0.622 & \cellcolor{top2color}0.567& \cellcolor{top2color}0.729 & \cellcolor{top2color}0.463 & \cellcolor{top2color}0.595 \\
MRFS (CVPR'24)       & 0.3832 & 0.1790 & 0.3268 & 2.4254 & 0.562 & 0.383 & 0.676 & 0.263 & 0.471 \\
LRRNet (TPAMI'23)    & 0.2490 & 0.1937 & 0.2848 & 2.1306 & 0.568 & 0.400 & 0.666 & 0.245 & 0.470 \\
DDFM (ICCV'23)       & 0.4549 & 0.2515 & 0.3829 & 2.5912 & 0.597 & 0.487 & 0.705 & 0.363 & 0.538 \\
TarDal (CVPR'22)     & 0.3992 & 0.2160 & 0.3576 & 2.7220 & 0.578 & 0.451 & 0.694 & 0.329 & 0.513 \\
ReCoNet (ECCV'22)     & 0.4129 & 0.1982 & 0.3395 & 2.3772& 0.581& 0.435 & 0.701 & 0.289& 0.501  \\
\midrule
\textbf{Ours}        & \cellcolor{top1color}\textbf{0.6282} & 0.2655 & \cellcolor{top1color}\textbf{0.5269} & \cellcolor{top2color}2.9998 & \cellcolor{top1color}\textbf{0.632} & \cellcolor{top1color}\textbf{0.576} & \cellcolor{top1color}\textbf{0.740} & \cellcolor{top1color}\textbf{0.469} & \cellcolor{top1color}\textbf{0.604} \\
\bottomrule
\end{tabular}
\end{adjustbox}
\end{table*}

\subsection{Robustness and Out-of-Distribution (OOD) Verification}

Prior-guided IVIF frameworks are often sensitive to prior quality, risking performance degradation in degraded scenes. To verify the resilience of $P^2$Fusion, we evaluate it under challenging scenarios like extreme overexposure and smoke occlusion (Figs.~\ref{fig:fig3}, \ref{fig:fig6}). Our framework mitigates prior dependency via the Teach-to-Fuse paradigm and its error-correction capability, as evidenced by SOTA results on M3FD and RoadScene without domain-specific fine-tuning (Table~\ref{tab:1}).

To test adaptability, we performed rigorous out-of-distribution (OOD) verification on the \textbf{DroneVehicle} dataset. This benchmark introduces a severe domain shift with aerial perspectives and severe visibility degradations. As summarized in Table~\ref{tab:ood}, P²Fusion consistently maintains superiority without any prior domain exposure during training. Specifically, in image fusion, our method achieves SOTA on two metrics and second place on another among five primary indicators. More importantly, in downstream object detection, P²Fusion achieves a comprehensive SOTA across all categories, outperforming the second-best competitor by 0.9\% in mAP$_{50-95}$. By dynamically mediating between intrinsic prompts and raw features, P²Fusion exhibits exceptional self-adaptability in unseen, hostile environments.


\section{Conclusion}
In this paper, we rethink the role of prior knowledge in infrared-visible image fusion and propose P²Fusion, a novel framework that shifts the fusion paradigm from static prior constraints to dynamic intrinsic prompting. By introducing a Teach-to-Fuse mechanism, we successfully distill task-specific physical priors—thermal saliency and spatial quality—into learnable regulators that procedurally govern the fusion process. Central to our architecture is the Gated Dynamic Expert Recalibration (GDER) module, which adaptively mediates modal competition and rectifies feature conflicts through expert specialization.

Extensive evaluations across five benchmarks and various challenging scenarios—including overexposure, smoke occlusion, and rigorous out-of-distribution (OOD) verification demonstrate that \text{P\textsuperscript{2}Fusion} not only achieves superior visual fidelity but also significantly enhances downstream perception. The remarkable generalization capability in unseen domains further underscores that our method effectively mitigates the inherent prior-dependency of traditional frameworks. Ultimately, this work provides a robust and versatile solution for multi-modal information integration, bridging the gap between high-level physical guidance and adaptive feature representation.

\section*{Acknowledgements}
This work was supported by the National Natural Science Foundation of China under Grant 62293543 and Grant 62322605.
\clearpage  

%
%
\bibliographystyle{splncs04}
\bibliography{main}
\end{document}